\documentclass{article}

\usepackage{amsmath,amsfonts,amssymb,times,graphicx,natbib,algorithm,algorithmic,hyperref}

\usepackage[accepted]{whi2018}
\usepackage{microtype}
\usepackage{graphicx}
\usepackage{subfigure}
\usepackage{booktabs}
\usepackage{todonotes}
\usepackage{hyperref}
\usepackage{amsfonts,nicefrac,microtype,amsmath,amssymb,mathtools}

\newcommand{\inR}[2]{\in \mathbb{R}^{#1 \times #2}}
\newcommand{\train}[1]{\mathbf{#1}_{\text{train}}}
\newcommand{\test}[1]{\mathbf{#1}_{\text{test}}}

\newcommand{\calf}{\mathcal{F}}
\newcommand{\dx}{\boldsymbol{\delta}}

\icmltitlerunning{Learning Qualitatively Diverse and Interpretable Rules for Classification}

\begin{document}

\twocolumn[
\icmltitle{Learning Qualitatively Diverse and Interpretable Rules for Classification}


\icmlsetsymbol{equal}{*}

\begin{icmlauthorlist}
\icmlauthor{Andrew Slavin Ross}{har,equal}
\icmlauthor{Weiwei Pan}{har,equal}
\icmlauthor{Finale Doshi-Velez}{har}
\end{icmlauthorlist}

\icmlaffiliation{har}{Paulson School of Engineering and Applied Sciences, Harvard University, Cambridge, MA 02138, USA}
\icmlcorrespondingauthor{Andrew Ross}{andrew\_ross@g.harvard.edu}

\icmlkeywords{interpretability, transparency}

\vskip 0.3in
]


\printAffiliationsAndNotice{}

\begin{abstract}
  There has been growing interest in developing accurate models that can also
  be explained to humans.  Unfortunately, if there exist multiple distinct but
  accurate models for some dataset, current machine learning methods are
  unlikely to find them: standard techniques will likely recover a complex
  model that combines them.  In this work, we introduce a way to identify a
  maximal set of distinct but accurate models for a dataset.  We demonstrate
  empirically that, in situations where the data supports multiple accurate
  classifiers, we tend to recover simpler, more interpretable classifiers
  rather than more complex ones.
\end{abstract}

\section{Introduction}
When building machine learning systems for high-risk situations with incomplete
information (e.g. healthcare), explanation is an important safeguard against
non-causal or otherwise nonsensical predictions
\citep{caruana2015intelligible}.  This observation has motivated a large body
of work in interpretable machine learning.  Much of that work falls into two
main categories: techniques for explaining existing models
\citep{craven1996extracting,lime,olah2017feature} and the construction of new
types of models that are inherently more interpretable
\citep{lou2012intelligible,decisionsets}.

In this work, we focus on interpretability specifically in contexts where the
data supports multiple functions of equal predictive accuracy for classification.  
In such situations, we demonstrate empirically that standard machine learning
techniques tend to recover combinations and conflations of the multiple functions,
regardless of the choice of model class.
While the individual functions each may be more or less interpretable, one can
certainly argue that combinations of functions---however complex they are individually---are
likely to be harder for humans to understand than just one of the functions
alone.  Providing these multiple options can help human experts choose one that
is likely to generalize best.

Our first contribution is a formal definition for a maximal set of diverse
classifiers.  Next, we propose an efficient method for training an ensemble of
diverse classifiers, which make accurate predictions on the training set but
whose outputs are statistically independent under small perturbations of their
inputs.  Empirically, we observe that our approach appears to separate out the
true underlying collections of (often human-interpretable) functions. Importantly, our 
definitions and training methods apply broadly across model classes.
Although the task of exploring the space of equally predictive models have been 
studied in literature, especially for real life applications \citep{liu2017multiple}, our
independence-based proxy, focus on maximal sets, and training methods are novel.

\section{Related Work}\label{sec:background}

One common proxy for interpretability is the minimality of explanatory factors
in a model, often quantified as sparsity with respect to dependence on input
variables. Minimal models in this sense can be obtained by feature selection,
that is, the task of finding classification functions that depend on as few
input variables as possible \citep{lasso,zou2005regularization}. In this
context, there is existing work that provides ways of traversing
the solution space of sparse models
\citep{thompson1978selection,habbema1977selection,jain2000statistical,hara2017enumerate}. However, these algorithms either do not explicitly
encourage diversity or have not been applied to deep models; neural network-specific feature selection methods \citep{verikas2002feature,kim2016opening,7280626,li2015deep,wang2014attentional} generally focus on producing a
single set of relevant input parameters.  Unlike these works, we do not focus
on the problem of specifically finding an interpretable model (or even a
collection of specifically interpretable models).  Rather, we focus on making
sure that we do not return models that could have been broken down into simpler
components. Our method often leads to the recovery of sparse solutions when
they exist, but generalizes to cases when these components are functions
of multiple input variables.

Our work is also related to ensemble creations methods that encourage
diversity during training. Typically, these methods obtain diverse ensembles by
training models on different datasets (or different weightings of the
same dataset) as in boosting \citep{freund1999short} and bagging
\citep{breiman1996bagging}, or by explicitly incorporating a term in the loss
function that encourages diversity of training predictions, as in Negative Correlation Learning
\citep{liu1999ensemble}.  However, ensemble methods that rely on maximizing
predictive differences on the training set are intended to improve the
generalization of a \textit{single} model created from the ensemble and do not
address the case where there are predictively equivalent models for the training
distribution we wish to isolate.  Our method instead focuses on training an
ensemble of models that all make the \textit{same} predictions on the training
set, but that generalize in qualitatively different ways outside it.

In this respect, our approach is similar to the ``find-another-explanation''
technique from \citet{rrr}, which is an iterative technique that can recover
models which are \textit{locally} sensitive to disjoint sets of features.
However, when normal training returns a model that is already globally
sensitive to all features, ``find-another-explanation'' terminates after one
iteration, even if that model could be decomposed into many others. Our
method resolves these problems by simultaneously training models and
quantifying diversity via local independence, a more flexible condition than local feature disjointness.

\section{Conceptual Framework}\label{sec:framework}

For simplicity, we will consider only the problem of binary classification. We
assume that our input space $\Omega$ is a subset of $\mathbb{R}^D$, and
represent the class of an input vector $x\in \Omega$ by a binary label $y\in
\{0,1\}$. Given a set of training inputs $\train{X} \inR{N}{D}$ and labels
$\train{Y} \in \{0,1\}^{N}$, we aim to learn a classification function $f_\theta\colon\Omega\to[0,1]$
(parameterized by $\theta$) that outputs a label probability.
We learn $f_\theta$ by minimizing a loss function $\mathcal{L}(\theta)$, which we shall  
take to be the cross-entropy, or the negative log-likelihood of the
observed data: $\mathcal{L}(\theta|x,y) = \mathbb{E}_{x,y}\left[ -y\log
f_\theta(x) - (1-y) \log (1-f_\theta(x)) \right]$ (which we empirically
approximate over batches of our dataset). Given $f_\theta$, the predicted label
$\hat{y}$ is computed from $f_\theta(x)$ by rounding to 0 or 1.

\subsection{Defining Maximal Sets}\label{sec:defn}
\begin{figure}
    \centering
    \includegraphics[width=0.43\textwidth]{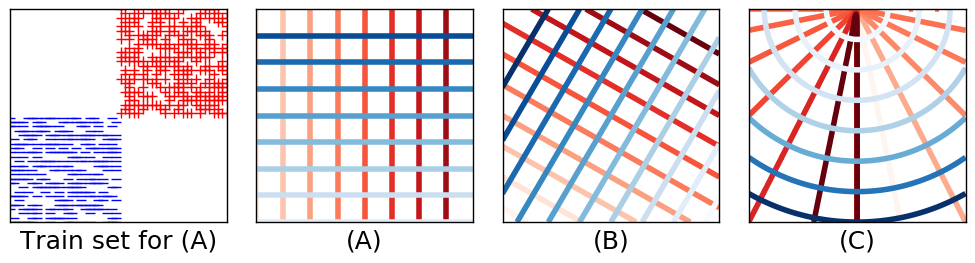}
    \caption{Illustrations of global and local independence. (A) shows level
    curves of two classification probability functions that are predictively
  equivalent on a training set (left), but make statistically independent
  predictions over $\Omega = \mathbb{R}^2$. (B) and (C) show pairs of classification
probability functions dependent on both input
variables that have orthogonal gradients everywhere, but are only \textit{locally independent} over
$\Omega = [0,1]^2$. However, we could make them globally independent by
changing the shape of $\Omega$ appropriately; e.g. for (B), by taking $\Omega$ to be a properly rotated rectangle, and for (C), by taking $\Omega$ to be a certain section of a unit disk.}
    \label{fig:indep}
\end{figure}

In this work, we study the case where our training data is generated from a
process with confounding factors, the result of which is that multiple
reasonable classifying functions can be fitted to the data. Our goal is to learn a maximal
large and maximally diverse set $\mathcal{F}$ of functions, each of which cannot be further
`decomposed' into combinations of other functions.
Formally, let $\mathcal{F} = \{f_1,\hdots,f_M\}$ be a set of classification functions, such that for each function $f_m\colon\mathbb{R}^{D} \to [0,1]$, the
accuracy rate of the predicted labels is greater than $1-\epsilon$ on the
training data for a fixed $\epsilon>0$. We call the set $\mathcal{F}$
\emph{independent} if the outputs of any pair of functions in $\mathcal{F}$ are
independent over the input space: $p(f_i(\mathbf{x}), f_j(\mathbf{x})) = p(f_i(\mathbf{x}))p(f_j(\mathbf{x}))$,
for $\mathbf{x} \sim \mathrm{Uniform}(\Omega)$ and $f_i, f_j \in \mathcal{F}$. Intuitively, we expect functions in
an independent set $\mathcal{F}$ to describe different ``hypotheses'' for how
inputs imply labels, and thus will generalize in qualitatively different ways
outside of our training dataset (which may only be supported on a subset of
$\Omega$). Two examples of an independent set of functions are shown in
Figure \ref{fig:indep}. Note that the two functions in Figure \ref{fig:indep}
(A) are supported on disjoint sets of input variables and hence can be
hypothetically recovered through processes of variable selection, while the
functions in (B) cannot be, since each depends nontrivially on
both input dimensions.

The requirement in the above definition that the outputs of two functions,
$f_i$ and $f_j$, be independent over the entire input space $\Omega$ ties our
definition to the shape of $\Omega$ (see Figure \ref{fig:indep}). However, for
interpretability, we argue that we are often more interested in functional
forms (i.e. functions as rules describing decisions) rather function values.
For this reason, we define a set $\mathcal{F}$ of functions to be
\emph{locally independent} if the outputs of every pair of functions in
$\mathcal{F}$ are independent for infinitesimally small isotropic Gaussian
perturbations of the input at every point in $\Omega$. Formally, we say that $f_i$ and $f_j$ are \emph{locally independent
at $x$}, if $p(f_i(x+\dx),f_j(x+\dx))= p(f_i(x+\dx))p(f_j(x+\dx))$, for $\dx \sim \mathcal{N}(0, \sigma^2
\mathbb{I})$ as $\sigma \to 0$.

Local independence admits an intuitive geometric interpretation.
For arbitrarily small $\sigma$, each $f$
is well approximated by its linearization and hence $f(x+\dx) \approx f(x) + \dx^\intercal\nabla_x f(x)$.
Therefore, the linearization of $f_i(x+\dx)$ and $f_j(x+\dx)$ are independent if and only if $\dx_i
\equiv \dx^\intercal\nabla_x f_i(x)$ and $\dx_j \equiv \dx^\intercal\nabla_x f_j(x)$ are independent. Since $\dx_i$ and $\dx_j$ represent 1D marginal slices
of an isotropic Gaussian along each gradient vector, $\text{Corr}(\dx_i,\dx_j) = \cos\left(\nabla_x f_i(x), \nabla_x f_j(x)\right)$, and
their mutual information
\begin{equation}
  \begin{split}
    I(\dx_i;\,\dx_j) &  = -\frac{1}{2}\ln\left(1-\cos^2\left(\nabla_x f_i(x), \nabla_x f_j(x)\right)\right)\\
                     & \approx I(f_i(x+\dx);\, f_j(x+\dx)).   \label{eqn:orthogonal_grads} 
  \end{split}
\end{equation}
The above equation relates statistical independence to
geometric orthogonality: $f_i(x+\dx)$ and $f_j(x+\dx)$ are independent in the limit as $\sigma \to 0$
if and only if their mutual information is 0, which occurs if and only if their input
gradients are orthogonal at $x$.  We call $f_i$ and $f_j$ \emph{locally independent}
without qualification if this condition holds for every point $x\in\Omega$.

Finally, we call a set $\mathcal{F}$ \emph{complete} if there does not exist
a classification function $f$, with accuracy greater than $1-\epsilon$ on the
training data, that is locally independent from the functions already in
$\mathcal{F}$. If $\mathcal{F}$ is a maximally large set that is both
locally independent and complete, we call it a \emph{maximal set} for
$(\train{X}, \train{Y})$. 

Intuitively, a maximal set captures the idea of an ensemble of models that are
equivalently predictive on the training set but offer qualitatively different
interpretations of the data. For example, in a feature selection problem, a
maximal set might contain functions that depend on disjoint sets of features.
In an image recognition problem, a maximal set might contain one function
sensitive only to the true object being detected and others sensitive to
associated but distinct contexts (though obtaining this set for image models in
practice may only be possible in latent representations).

We claim that the local independence property of maximal sets produces
functions that generalize differently with respect to changes in the input
distribution while the maximality condition encourages functional simplicity
(in the sense that each function in a maximal set cannot be further decomposed
into a combination of locally independent functions). In Section
\ref{sec:results}, we offer evidence in support of our claims in experiments
using synthetic data, and in Section \ref{sec:future} we discuss a broad basis
of theoretical and empirical support for these claims as future work. Finally,
we note that while the classification probability functions $f_\theta$ are
represented by neural networks in this paper, our analysis is not specific to
this model class.

\subsection{Learning Maximal Sets}\label{sec:training}
In general, learning a maximal set $\mathcal{F}$ for a training set requires
learning the size, $M$, of $\mathcal{F}$. While one can develop algorithms to
iteratively build up $\mathcal{F}$, in this paper, we suppose that the size of
$\mathcal{F}$ is known. This assumption is often reasonable in practice, since
$M$ is bounded by the input dimension $D$, and, empirically, we find that for a reasonable range of
settings of parameters in our training objective function, setting $M$ too high causes local independence training to return models with low training accuracy rather than similar gradients (so the procedure can be terminated then).

Fixing $M$, we propose a simple optimization procedure (Equation
\ref{eqn:learning_obj}) to jointly learn the optimal parameters,
$\theta_1^*,\ldots,\theta_M^*$, of models (chosen from any model class) that
approximate functions in a maximal set. We denote the set of $M$ models by
$\widehat{\mathcal{F}} =\{ \widehat{f}_{\theta_1^*}, \ldots,
\widehat{f}_{\theta_M^*}\}$. In our training objective, we maximize the
predictive value of the models while penalizing the local dependence the
probability functions they represent over the training data. We call this
training procedure \emph{\textbf{local independence training}}.
\begin{equation}
\begin{split}
  \theta_1^*,&\ldots,\theta_M^* = \underset{{\theta_1,\ldots,\theta_M}}{\mathrm{min}}\; \mathbb{E}_{x,y}\left[
  \sum_{m=1}^M \mathcal{L}(\theta_m|x,y)\right.\\
     &+ \left.\lambda \sum_{a=1}^M \sum_{b=a+1}^M
\cos^2\left( \nabla_x \widehat{f}_{\theta_a}(x), \nabla_x \widehat{f}_{\theta_b}(x) \right)\right].
    \end{split}\label{eqn:learning_obj}
\end{equation}
The first term in Equation \ref{eqn:learning_obj} measures the predictive value
of the functions in $\widehat{\mathcal{F}} $ as the sum of our single-model
loss functions. The second term penalizes
non-orthogonal gradients using their squared cosine similarity
$\cos^2(v,w) = \frac{(v^\intercal w)^2}{(v^\intercal v)(w^\intercal w)+\epsilon}$ (here we add
$\epsilon=10^{-6}$ to the denominator for numerical stability), which promotes
local independence. In
practice, we take gradients of the log-odds rather than the probability to
prevent underflow. $\lambda$ specifies the penalty strength, and should be
chosen to make the penalty term (which is bounded by $M$ and $\lambda$)
comparable in initial magnitude to the sum of cross-entropies.

\section{Experimental Results}\label{sec:results}

In the following we present results from three sets of experiments with
ground-truth maximal sets. We show that local independence training produces
sets of models that recover them, while normal training across a variety of
model classes does not.  Code and data to reproduce these experiments will be
made available at \url{https://github.com/dtak/local-independence-public}.

\subsection{2D Illustrative Examples}
\begin{figure}[H]
    \centering
    \includegraphics[width=0.33\textwidth]{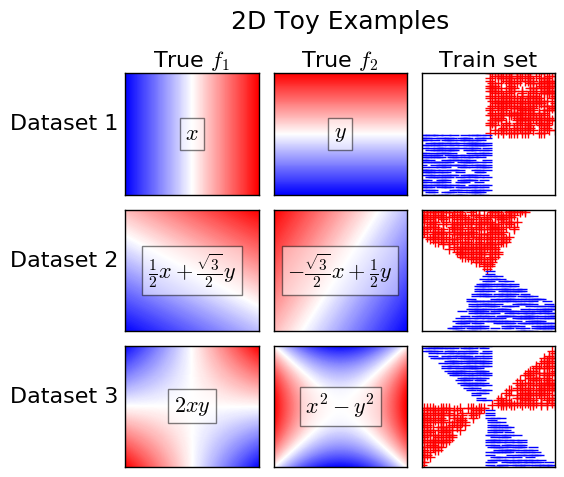}
    \caption{Three 2D training datasets each created by taking points from
    $\Omega$ where functions $f_1$ and $f_2$ have the same sign, hence yielding
  the same labels.}
    \label{fig:synthetic_setup}
\end{figure}

As an initial experiment, we construct three 2D datasets over the input domain $\Omega = [-10,10]^2$ (see Figure \ref{fig:synthetic_setup}). The training inputs $\train{X}$ are taken from a subset of $\Omega$ where, respectively, the three pairs of functions
\begin{align}
\calf_1 &= \{x,y\},\\
\calf_2 &= \left\{\frac{1}{2}x+\frac{\sqrt{3}}{2}y,\,\,-\frac{\sqrt{3}}{2}x+\frac{1}{2}y\right\}\\
\calf_3 &= \{2xy, x^2-y^2\}
\end{align}
have the same sign (and hence yield the same labels $\train{Y}$). Although we train models
over data from this restricted domain, we evaluate them over the entirety of $\Omega$ and test their agreement with $f_1$ and $f_2$.
As our model class, we use 256x256 multilayer perceptrons with softplus activations, trained with Adam at a learning rate of 0.001. For this and all other experiments, we use $\lambda=0.1$.

\begin{figure}[H]
    \centering
    \includegraphics[width=0.48\textwidth]{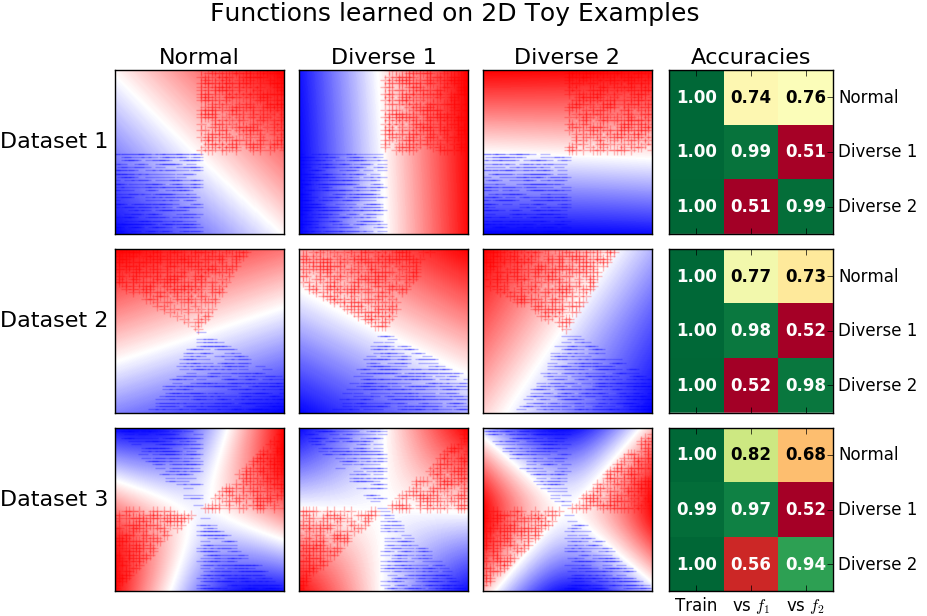}
    \caption{Comparison of classifiers obtained using normal training and local
    independence training on the datasets in Figure \ref{fig:synthetic_setup}.
  For each dataset, local independence training yields a set of models (labeled
Diverse 1 and Diverse 2, with contour plots showing log-odds model outputs)
that recovers the ground truth functions, whereas normal training recovers a
dense combination (which is reflected in accuracy results vs. the true $f_1$
and $f_2$ over all of $\Omega$).}
    \label{fig:synthetic_result}
\end{figure}

In Figure \ref{fig:synthetic_result}, we show that the models produced by local independence training recovers
these ground truth functions fairly well, while normal models learn a dense combination.

Although these examples are simple, they pose challenges for many ensemble creation approaches referenced in Section \ref{sec:background}. For example, the ``find-another-explanation'' method from \citet{rrr} is unable to recover ground-truth functions even on the first case, because the initial model learns to assign importance to both features everywhere.
Feature selection and enumeration methods such as \citet{hara2017enumerate} cannot handle the
second and third cases, where true functions depend on multiple input features.

\subsection{8D Feature Selection Example}

\begin{figure}
    \centering
    \includegraphics[width=0.33\textwidth]{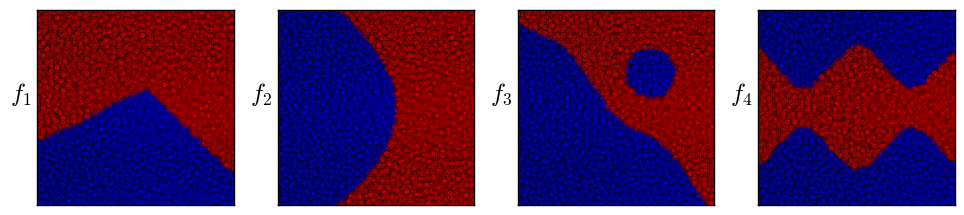}
    \caption{Toy 8D dataset is generated by sampling from the region in $\mathbb{R}^8$ where four classification probability functions agree. Each function depends on a unique set of two axes; we show their decision boundaries here.}
    \label{fig:toy8d}
\end{figure}

\begin{figure}
  \hspace{0.15cm}\includegraphics[width=0.39\textwidth]{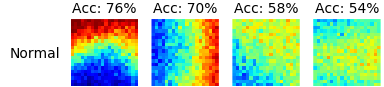}\includegraphics[width=0.04\textwidth]{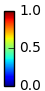}\\
   \includegraphics[width=0.40\textwidth]{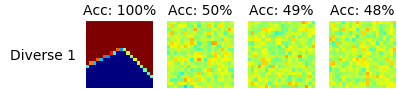}
    \includegraphics[width=0.40\textwidth]{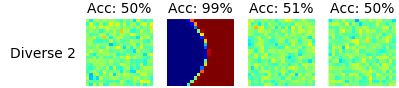}
    \includegraphics[width=0.40\textwidth]{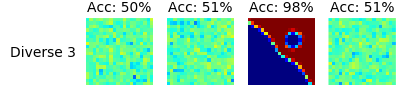}
    \includegraphics[width=0.40\textwidth]{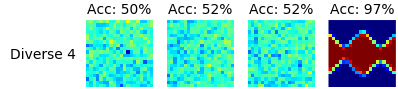}
    \caption{Accuracy on each test set and 2D projections of mean classification probability are shown. Local independence training recovers the functions in our ground-truth $\mathcal{F}$ while a normally trained model learns a dense combination of them.}
    \label{fig:toy8d_results}
\end{figure}

As a larger example, we consider an 8D dataset where the labels of training
data on a subset of $\Omega=[-20,20]^8$ can be redundantly determined by any
one of four functions of two non-overlapping dimensions (Figure
\ref{fig:toy8d}). We again train 256x256 MLPs, with ReLU activations, and we reduce the learning rate from 0.001 to 0.0001 for local independence training.
On this dataset, we find that even ``interpretable'' model classes
like logistic regression and decision trees learn functions that are dense combinations of all
the ground truth functions (see Table \ref{table:non-nn-accuracies}). The same is true for neural networks.
However, in Figure \ref{fig:toy8d_results}, we see that local independence training encourages
an ensemble of neural networks to learn functions that correspond very closely to
the ground truth $\mathcal{F}$ and make independent predictions over $\Omega$.

\begin{table}
  \centering
  \small\begin{tabular}{|c|c|c|c|c|c|}\hline
    \textbf{Model}  & \textbf{Train} & \textbf{Test 1} & \textbf{Test 2} & \textbf{Test 3} & \textbf{Test 4} \\\hline
    Logistic Reg. & 0.99 & 0.70   & 0.75   & 0.61   & 0.49   \\\hline
    Decision Tree & 1.00 & 0.57   & 0.77   & 0.54   & 0.57   \\\hline
    Rand. Forest  & 1.00 & 0.67   & 0.72   & 0.56   & 0.50   \\\hline
    SVM           & 1.00 & 0.47   & 0.55   & 0.59   & 0.65   \\\hline
  \end{tabular}
  \caption{Function-specific accuracies for four non-neural model classes (trained on the confounded Toy 8D dataset). All models learn a dense combination of almost all functions, even ``interpretable'' logistic regression and decision tree classifiers.}
  \label{table:non-nn-accuracies}
\end{table}

\subsection{Latent Space Example}

\begin{figure}[htb]
    \centering
    \includegraphics[width=0.3\textwidth]{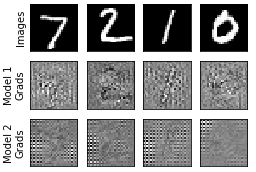}
    \caption{Pathological behavior of convolutional networks trained with a local independence penalty on MNIST. When the input space is high-dimensional with ``slack'' components that never change, orthogonality can be achieved trivially by assigning infinite gradient magnitudes to them.}
    \label{fig:pathological}
\end{figure}

As a final experiment, we consider learning diverse image classifiers for
dSprites \citep{dsprites17}, which is a dataset of $64\times64$ images
generated from five independent latent factors (shape, scale, rotation, and
$x$- and $y$-position). In this case, $\Omega$ is defined in the latent space
of the data's true generative factors. Image classification problems pose a
challenge for local independence training in input space, because images are
very high-dimensional and sometimes contain features that do not vary
significantly for any input examples. When such ``slack'' features exist, we
can trivially minimize Equation \ref{eqn:learning_obj} by assigning very large
gradient values to those components (see Figure \ref{fig:pathological}), which does not hurt
training accuracy because those features never change. Another pathological
solution can occur when a set of models ``divvy up'' the feature space and
simply consider alternate pixels, which leads to orthogonal input gradients but
not qualitative diversity. The crux of this problem lies in the fact that we would like
our classification functions to be locally independent on the data manifold, not $\mathbb{R}^D$.

To resolve these challenges, we suggest applying local independence training after a preprocessing step
where we learn a lower-dimensional latent representation, e.g. using an autoencoder, which can often learn an approximation of the true data manifold.
After fixing the encoder weights, we can apply local independence training with latent space input gradients.
If the low-dimensional representation already \textit{disentangles} \cite{bengio2013deep,higgins2016beta} latent factors that correspond to different confounding rules, then the problem of learning a maximal set reduces to one of feature selection, which our method can solve. However, even if the
representation only disentangles these rules up to a nonlinear conformal
transformation, local independence training should still be able to recover
them.

\begin{figure}[htb]
    \centering
    \includegraphics[width=0.33\textwidth]{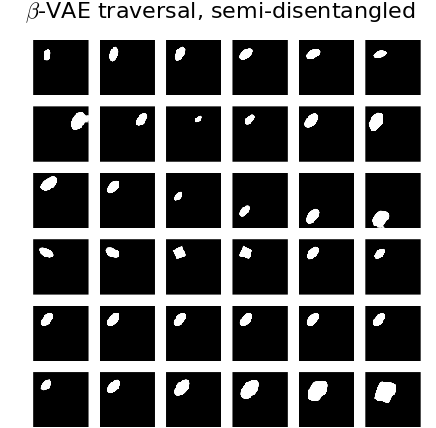}
    \caption{Latent traversal of the $\beta$-VAE we trained, showing all dimensions with variance less than 0.7. Although $x$- and $y$-position appear to be disentangled from each other, both are entangled with scale (which itself appears entangled with shape). }
    \label{fig:beta-vae}
\end{figure}

For our experiment, we trained a convolutional $\beta$-VAE using the updated formulation of \cite{burgess2018understanding,miyosuda} on the full dSprites dataset, using $C=20$ and $\gamma=1$. At these settings, we achieve partial but not complete disentanglement of the latent generative factors, which can be seen in Figure \ref{fig:beta-vae}. For classification, we construct $\train{X}$ by selecting points whose scale, $x$-, and $y$-positions are all above or all below the median values for those generative features (excluding about $\frac{7}{8}$ of the dataset). The training labels are all set to 0 or 1 for points with factors below or above all three medians -- so for our ground-truth maximal set,$\mathcal{F}=\{f_1,f_2,f_3\}$, we define $f_1$, $f_2$, and $f_3$ to indicate whether scale, $x$-, and $y$-position are above their respective medians. The inputs for testing, $\test{X}$, are a held-out subset of 10000 uniformly distributed examples from the full dataset. This setup mimics the scenario where our unlabeled data (used by the autoencoders to learn the latent representation) is more diverse than our labeled data, which is collected with selection bias. We freeze the weights of the $\beta$-VAEs after training and learn the same 256x256 MLP classifier on top of their latent representations.


\begin{figure}[htb]
  \centering
  \includegraphics[width=0.4\textwidth]{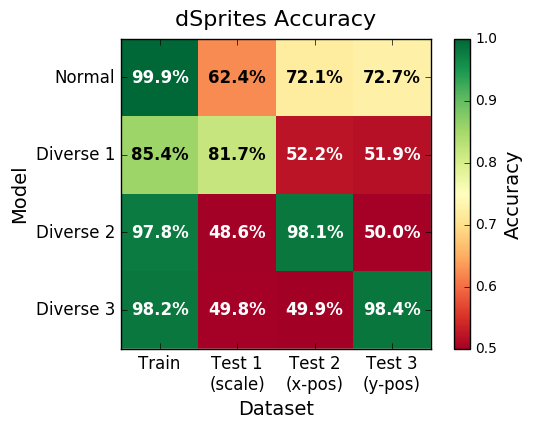}
  \caption{Accuracy results on dSprites for MLP classifiers trained on representation from our $\beta$-VAE. Normal training learns a classifier dependent on multiple latent factors and does better than random guessing but worse than optimally on test sets generated using only one latent factor. Local independence training recovers a set of models (Diverse 1, 2, 3) that each depend on only one latent factor.}
  \label{fig:dsprites-results}
\end{figure}

\begin{figure}[htb]
    \centering
    \includegraphics[width=0.4\textwidth]{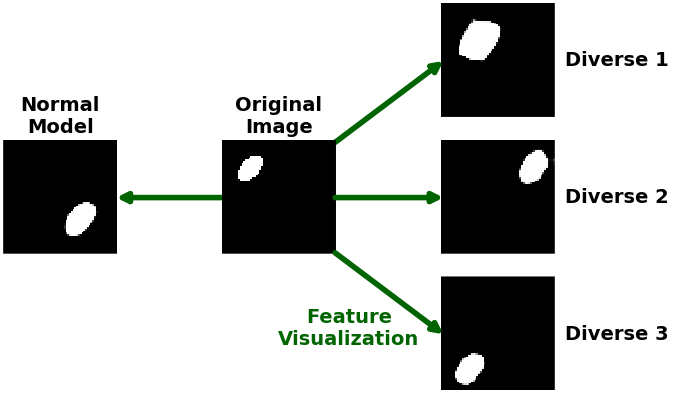}
    \caption{Feature visualization of the normally trained model (which exhibits sensitivity to all latent factors) compared to the three models obtained from local independence training (each of which is mostly only sensitive to one factor).}
    \label{fig:dsprites-models}
\end{figure}

As we can see in Figure \ref{fig:dsprites-results}, classifiers trained normally on this dataset learn a dense combination of all three functions in $\mathcal{F}$. With local independence training, we are able to separate out $f_2$ and $f_3$ (the $x$- and $y$-position-based functions) almost perfectly. We have a more difficult time separating out scale, which is entangled with $x$- and $y$-position in our representation, but we do achieve significantly higher accuracy on the scale-based test set for the first model in our ensemble. Interestingly, this model also achieves imperfect training set accuracy, which indicates that local independence training can find solutions that smoothly trade accuracy for independence when both are not achievable. In Figure \ref{fig:dsprites-models}, examining the models more closely via feature visualization \cite{olah2017feature} in latent space, we see that each model in our diverse ensemble seems to correspond fairly closely to the independent generative factors. By learning an ensemble of locally independent classifiers, we are actually able to achieve greater disentanglement of latent generative factors than our imperfect representation.

\section{Discussion \& Future Work}\label{sec:future}

In this paper, we demonstrated that standard ML training techniques---when faced with a data set that supports multiple confounding decision rules---learn complex combinations of those decision rules.  We suggested that a heuristic way to separate them out is to train a maximal ensemble of models whose corresponding classification functions are locally independent.  We designed an algorithm to learn a maximal set of locally independent models through optimizing for orthogonal gradients during training. In three sets of experiments on datasets engineered to support a set of functions with `simple' functional forms, we show that our training procedure recover these functions, whereas normally trained classifiers learn a dense, and hence more complex, combination of them.

There is ample opportunity for both empirical and theoretical future work. We introduced the formalism of a maximal set for describing the modeling ambiguity of a dataset, but there is room to improve this definition. Empirically, in this paper, we intentionally ran experiments where we had ground-truth knowledge about the number of functions supported by the data, the complexity of their functional forms, and the fact that all of them could be used to classify the dataset with perfect accuracy (though in separate experiments, we have found the method to be fairly robust to noise). However, future explorations of this idea should consider cases where the true number of underlying predictive functions is unknown and where some of them are only partially predictive of the label.  It would also be important to test whether these maximal sets are indeed more interpretable than their more complex counterparts via user studies, or whether they can be used as bases to construct functions that are.  


\paragraph{Acknowledgments}
FDV acknowledges support from AFOSR FA9550-17-1-0155. ASR and WP acknowledge support from the Institute for Applied Computational Science (IACS) at Harvard University. The authors thank Melanie Fernandez Pradier for helpful discussions.

\bibliography{bibliography}
\bibliographystyle{icml2018}

\end{document}